# A proposed Bi-LSTM method to fake news detection

Taminul Islam*
Department of Computer Science & Engineering
Daffodil International University
Dhaka, Bangladessh
taminul@ieee.org

MD Alamin Hosen
Department of Computer Science & Engineering
Daffodil International University
Dhaka, Bangladessh
alamin15-11132@diu.edu.bd

Akhi Mony
Department of Computer Science & Engineering
Daffodil International University
Dhaka, Bangladessh
akhi15-11130@diu.edu.bd

MD Touhid Hasan
Department of Computer Science & Engineering
Daffodil International University
Dhaka, Bangladessh
touhid15-10500@diu.edu.bd

Israt Jahan
Department of Computer Science & Engineering
Daffodil International University
Dhaka, Bangladesh
isratjahan.cse@diu.edu.bd

Arindom Kundu
Department of Computer Science & Engineering
Daffodil International University
Dhaka, Bangladessh
arindom15-10557@diu.edu.bd

*Abstract*— Recent years have seen an explosion in social media usage, allowing people to connect with others. Since the appearance of platforms such as Facebook and Twitter, such platforms influence how we speak, think, and behave. This problem negatively undermines confidence in content because of the existence of fake news. For instance, false news was a determining factor in influencing the outcome of the U.S. presidential election and other sites. Because this information is so harmful, it is essential to make sure we have the necessary tools to detect and resist it. We applied Bidirectional Long Short-Term Memory (Bi-LSTM) to determine if the news is false or real in order to showcase this study. A number of foreign websites and newspapers were used for data collection. After creating & running the model, the work achieved 84% model accuracy and 62.0 F1-macro scores with training data.

## I. INTRODUCTION

Social media now gathers and shares news as well as providing a constant supply of fresh information on the internet every day. The function of news is to inform people about current events across the world. Social media as a source of collecting news is growing increasingly popular over conventional media, such as television, radio, and newspapers, due to a number of factors. There are a variety of reasons for this. One is because it is simple to access, which results in reduced costs and less time required to get news. The more outrageous the story, the more likely it is to become viral on social media, which increases the likelihood of its being false news. Social media platforms like Facebook, Twitter, and Instagram open up new facilities with some obstacles like bogus news being shared, and plenty of social bots [1].

Fake news is deception or rumor provided to the people as if it were real information with the goal of influencing people's thinking and molding public opinion into political action, and fake news publications have often found ways to earn money online through their misinformation. Fraudulent news can dupe the public by seeming similar to legitimate websites or utilizing identical URLs to reputable news outlets. These websites may be found across the world, such as in the United States, China, Russia, and other nations. To a large extent, humans are solely responsible for fake news and political inconsistency, both of which serve selfish interests. There are already around 23 million social bots on Twitter, with 140 million on Facebook, and 27 million on Instagram. The spread of political division as a result of fake news is inarguable. The current US presidential election has brought attention to the issue of false news. It has become the home of posting bogus news on social media [2].

Approximately 6 in 10 adults in the United States use the internet, and about one in six uses Twitter [3]. claims that in the 2016 US presidential election, false news attracted people's attention and had an impact on people's opinions. It therefore became a global menace. In order to stop the spread of false news, it was important to design ways to detect it. The work introduced and also applied a machine learning approaches which is Bidirectional Long Short-Term Memory (Bi-LSTM). And we got the best accuracy on in this model.

## II. RELATED WORKS

After the U.S. presidential election in 2016, the machine learning research community began to dedicate a lot of time to finding fake news, which gained a lot of attention in the general media has done an extensive study on the identification of false news by leveraging information from the sources. They noted the following characteristics that are utilized in content-based classification: Lexical, Syntactic, Visual, Statistical, Users, Post, and Network. Deep neural network models were developed by [4] to identify false news. The dataset used in the experiments was real-world (Buzzfeed and PolitiFact). Their technique of classifying has been separated into three distinct sections.

The first two are text-based categorization systems; the other two utilize both social context and social network data. The model that performs best has been determined to be a deep neural network. To be certain that a news story was genuine, [5] looked at language characteristics to spot false news. The results of their study evaluated four language aspects. In this case, we see evidence from syntax, emotion, grammar, and readability.

The sequential neural network-based model improved on the prior machine learning based approach, outperforming it by over eight percent. The authors, [6] conducted a study that was content-based. Consider the number of unique words, the average number of words per post, and the average amount of characters each post to determine the average number of characters submitted each week. They implemented several machine learning methods and they managed to achieve the best results using SVM (Support Vector Machine). They obtained a result of 95.70%.

An ensemble model outperformed a single algorithm-based model, [7] examined. A capsule neural network was used by [8] to identify fake news. There are two distinct categories of news in their dataset. An infographic which has short texts and another that has medium to long words. The keywords "dynamic," "non-static," and "multichannel" have been used in word embedding techniques. [9] collect statistics on false news. The data contained in their dataset has several languages (English, Spanish, French, Portuguese, Hindi, Turkish, Italian, Chinese, Croatian, Telugu), thus their dataset offers a great opportunity for those who want many languages. In contrast to well-crafted artificial datasets, fake news datasets are often made up of data from just one language. They did a bench analysis of their dataset and they scored 76% with the BERT-based classifier.

The benchmark dataset for false news presented by [10] included fact-checked news articles from 92 fact-checking websites in many languages for COVID-19. In his paper, [11] outlined an annotation system for social media data collected in several modes. They've developed a semi-automated system for gathering multi-modal annotated social media data that combines the efforts of humans and robots. This framework has also been used to obtain false information on COVID-19.

The transformer model suggested by [12] includes human reasoning and metadata for improved performance in classifying false news inside a given domain dataset. To deal with diverse inputs, they employed several BERT models, each with shared weights. Fake news identification might be made easier with an integrated multi-task model, according to [13]. Considering that either one of the two might have a larger rate of false news, they looked at both subjects and writers. They're looking into how subject labels and other contextual information affect short-form false news.

### III. METHODOLOGY

There were six steps in the process of completing this task. A short representation of these steps may be seen in this paragraph. Collecting data from many online news sources is the first stage in constructing a complete dataset. The next step is to preprocessing the data. After that, use Bi-LSTM classifiers to classify the dataset and build the model. It is then possible to measure the model's performance using a variety of measures like accuracy, recall and precision once it has been trained and tested on the dataset.

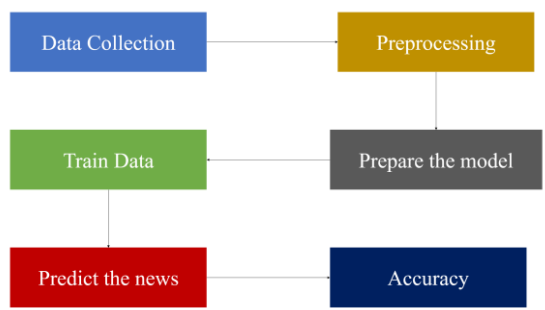

Fig. 1 Proposed model workflow

The most important aspect of research is the subject matter and the field in which it is being conducted. Our study is primarily news-based. When it comes to Research, it's all about what you're studying. We also digest the news and offer it to our audience in a unique way. It's not enough to convey a notion of what research is all about; it has to be explained in detail. Consequently, in order to have a comprehensive understanding of the subject, you must be familiar with its several subfields. For the sake of this study, we've covered all of the relevant topics. Researchers employ a variety of tools to help them conduct their studies, including a wide range of instruments. Fig 2 shows the proposed flowchart of this work.

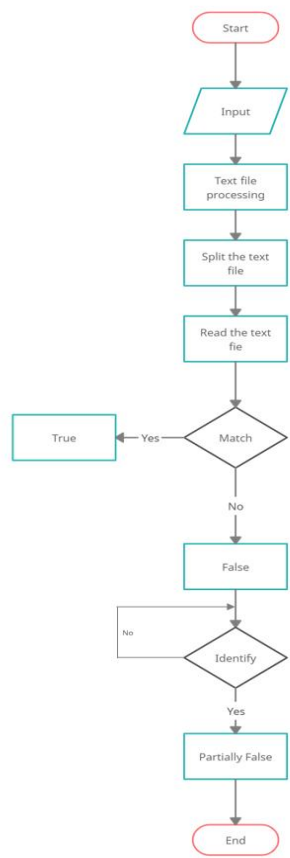

Fig. 2 Proposed flowchart of this work.

### A. Data Description

Data is one of the most important aspects of any research project, and without it, it is nearly impossible to conduct a study. The results of research are utilized for a variety of testing reasons. And, armed with this information, we can proceed forward with our investigations. We have collected data from various website. We have collected several kinds of news from different international websites.

Fake news and rumor are the biggest hazards to society in this digital age since they may have a wide range of harmful effects. It's quite difficult for readers to tell the difference between true and bogus news. There are a number of online news portals, blogs, and sites that do not have the legal authority to publish news, but they continue to post rumors or useless news with sizzling headlines in order to attract visitors. Table 1 shows the quantity of data collections along with platforms.

TABLE 1: DATA QUANTITY WITH PLATFORMS

| Platform | Data |
|---|---|
| CNN | 550 |
| Fox News | 220 |
| The Guardian | 340 |
| American News | 670 |
| BVA News | 425 |
| The Buston Tribune | 410 |
| Fox-nees24.com | 362 |
| **Total** | **2977** |

We have collected total 2977 data from several websites. There is a variation of different data. The data was classified in to three categories; True, False & Partially false. Where True data contains 1712, False contains 856 & partially false contains 406 news data. Figure 3 shows the categories of data with size.

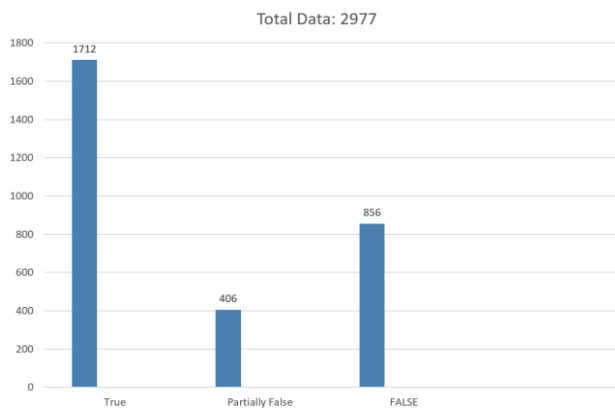

Fig. 3. Data categories with size

### B. Data Pre-Processing

When we're doing research, the first thing we do is preprocess data. The initial step in data mining is to process the data. We get news from a variety of sources for this purpose. In order to fix this, we begin by preparing the data. Each news outlet publishes a variety of stories, and these stories form a Beginning Data Set. These data sets are broken down into a variety of numbers. Processes this data on a one-by-one basis. Fig 4 shows the steps of the data preprocessing

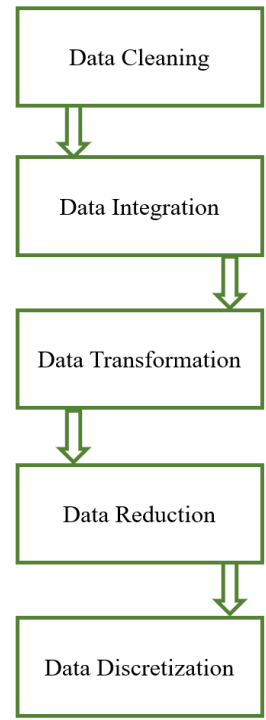

Fig. 4. Proposed data preprocessing method

*1) Removing puntuation:* There are a lot of punctuations, links, numerals, and special characters in news items that in most circumstances have no bearing on whether or not the news is accurate or false. Adding to this, punctuation appears often and has a significant influence on the metrics for punctuation, but has no effect on the categorization of the text.

*2) Capitalization:* Although it is possible to use upper and lowercase letters in a sentence, it's preferable to use the same register level in a computational model. When it comes to digits, it doesn't matter what sort of register level you employ. We utilized lowercase registers in this paper.

*3) Lemmatization:* Using lemmatization, the number of words with similar meanings may be reduced. Using either lemmatization or stemming, the entire form of a word is reduced to its root form. In contrast to lemmatization, stemming merely removes the beginning and end of a word to get its root. In this paper, we employed stemming to accomplish our goal.

*4) Removing Stop-words:* We can cut the number of words in our data even further without affecting the accuracy of our model. A stop-word is a term that appears repeatedly in a piece of writing but has no bearing on the meaning of the piece. Cleansing our data is now complete, and we are ready to begin putting it together into a bag of words [14]. In order to make the data trainable, we need to remove all of the text and strings from the data. For this reason, we transform all of our text into numbers so that it may be used as a learning tool. Our final trainable data is created by vectorizing our text data using the TF-IDF vectorization approach.

*C. Proposed Model Working Procedure*

CNN, Fox News, The Guardian, American News, BVA News, The Buston Tribune, and Fox-news24.com are the sources of our raw data. Our researcher collects our data manually. The news is saved in a text document file when it has been gathered. There are a lot of unnecessary elements in these files, such as an html tag name and a lot of unnecessary data. Since then, it has become clear that the data must be cleaned. That signifies that the raw data has been pre-processed in order to be used in the model.

*1) Bi-LSTM model:* Compared to Long Short-Term Memory, Bi-Directional long short-term memory (BiLSTM) is superior in sequence categorization (LSTM). The BiLSTM is made up of two LSTMS, one for forward input and the other for backward input [15]. Using concealed states, it is possible to exchange data in both directions. On every time step, two LSTMs' outputs are combined to form one. The BiLSTM approach helps to reduce the limitations of standard RNN. The context is better understood because to BiLSTM's high level of accuracy.

*2) Data Cleaning:* Our data pre-processing task is made easier by the usage of a script file. It is this python script that removes all html tag names. Make sure there are no superfluous gaps in the text. Take off all of the new lines from each news item and arrange it in a single line. Pre-define each news category by assigning an integer number. We can really receive all of our categorized news in separate files using this method, but the data that is returned is already pre-processed and classified..

*3) Excluded Words Removal:* Python code is used to categorize a news item into a certain category. Our system is now ready to develop a model after combining all of the news into a single file. This necessitates some preparatory work. We compile a list of English words that has nothing to do with the news category. Using the term Excluded words, we dubbed the list Excluded words list. Our input file is being checked to see whether any of the terms we've omitted are there. If it's there, it must be eliminated.

*D. Feature Extraction and fitting dataset*

Feature selection and extraction are the primary components of this approach's classifying step. When it comes to categorization, it really makes the final decision. We acquire enormous volumes of data. An approach is required in order to make sense of this data. They cannot be processed manually. In this case, feature extraction becomes relevant.

We split our dataset in two to begin the modeling process.
1.Training Dataset

2.Testing Dataset.

80% of the data will be used for training, and 20% will be used for testing. And this is what we anticipate to see in our model, as well We import the sklearn package because we are working with Bi-LSTM classifiers. This classifier is capable of generating an integer that reflects the predicted news category. The figure 4.3.1 illustrates the data ratio of this research work.

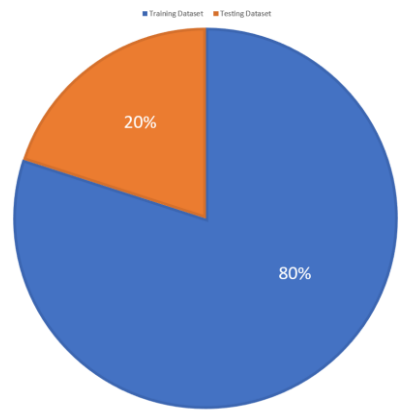

Fig. 5. Data Ratio

## IV. RESULT AND DISCUSSION

Our BiLSTM model was built with the help of the Keras library. Using a glove embedding of 100d, a model may be created This research use a sequential model as the basis for its analysis. A variety of techniques are employed, including embedding, dropout layers, and a layer with 256 neurons that is completely linked. We have a multiclass dataset.

Because of this, the output layer is applied using soft-max activation. As a result of this design, no data from the test set or the training dataset are ever used. The model is trained with 20 epochs and 128 batch sizes of training data. We found the accuracy of this model is 84%, and the F1-macro score is 62.0.

The suggested model's output reveals the identify of the news item that was shown. The news is real, fake, or half false, depending on who you ask. True is assumed to be 2, false is considered to be 0, and partially false is considered to be 1 for determining the identification of the news because the RNN model cannot operate with text. In order to train our model, we employed 20 epochs and obtained an accuracy of 84% and a f1 -macro average score of 62 %. The accuracy and f1 macro average score of the model are shown in the following table 2, which also contains information on the classification report.

TABLE 2: CLASSIFICATION REPORT OF THE MODEL

| Class | Precession | Recall | F1 score |
|---|---|---|---|
| True | 0.82 | 0.97 | 0.89 |
| False | 0.92 | 0.91 | 0.91 |
| Partially False | 0.50 | 0.03 | 0.06 |

The following figure also shows the accuracy and f1 macro average score of the model.

```
Evaluating Model ...
              precision    recall  f1-score   support

           0       0.92      0.91      0.91       174
           1       0.50      0.03      0.06        68
           2       0.82      0.97      0.89       353

    accuracy                           0.84       595
   macro avg       0.74      0.64      0.62       595
weighted avg       0.81      0.84      0.80       595
```

Fig. 6. Accuracy and f1 macro average score of the model

TABLE 3: ACCURACY REPORT

| Method | Accuracy | F1 macro average |
|---|---|---|
| Bi-LSTM | 84% | 62% |

The following diagram demonstrates the graphical representation of our suggested model's accuracy vs evaluation accuracy (Figure 4.3) and loss vs evaluation loss (Figure 4.4), which are both obtained by our proposed model. It is evident from those images that our proposed model is gaining knowledge from its previous.

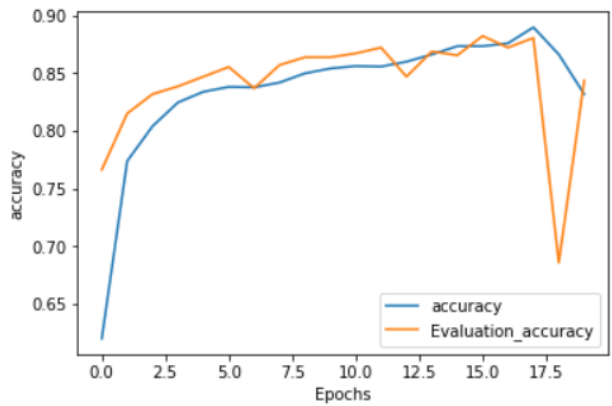

Fig. 7. Accuracy VS Evaluation Accuracy

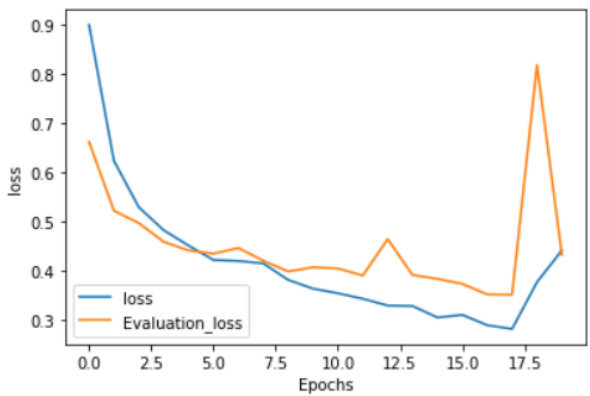

Fig. 8. Loss VS Evaluation Loss

## V. CONCLUSIONS

A BiLSTM model for identifying fake news was suggested in this study. Filter techniques were utilized to remove unnecessary attributes from our dataset in order to speed up the training process. In order to make the features machine understandable, count vectorizer and word tokenizer are utilized. Therefore, to reduce the time it takes to train our proposed model, features, stopwards, Stemming, and other data preparation activities are employed. This model has an overall accuracy rate of 84% and a macro score of 62.0, which is satisfactory. There are a few issues that need to be addressed. We can improve accuracy and performance by combining BiLSTM with CNN. In order to improve our model's performance, we must spend a lot of effort and money on high-quality hardware components. In the future, we want to use a larger, more diverse dataset to test our suggested approach and overcome our current constraints.

## VI. ACKNOWLEDGEMENT

This is an excellent chance for our team to contribute to this research endeavor. We would like to express our gratitude to Israt Jahan, a lecturer in the Department of Computer Science and Engineering at Daffodil International University, for helping us with our research. A special thank you goes out to Daffodil International University for their assistance in supplying us with essential materials for our project.